# Detection of Correlated Alarms Using Graph Embedding

Hossein Khaleghy
Department of Electrical and
Computer Engineering
Isfahan University of Technology
Isfahan 84156-83111, Iran
h.khaleghy@ec.iut.ac.ir

Iman Izadi
Department of Electrical and
Computer Engineering
Isfahan University of Technology
Isfahan 84156-83111, Iran
iman.izadi@iut.ac.ir

*Abstract*—**Industrial alarm systems have recently progressed considerably in terms of network complexity and the number of alarms. The increase in complexity and number of alarms presents challenges in these systems that decrease system efficiency and cause distrust of the operator, which might result in widespread damages. One contributing factor in alarm inefficiency is the correlated alarms. These alarms do not contain new information and only confuse the operator. This paper tries to present a novel method for detecting correlated alarms based on artificial intelligence methods to help the operator. The proposed method is based on graph embedding and alarm clustering, resulting in the detection of correlated alarms. To evaluate the proposed method, a case study is conducted on the well-known Tennessee-Eastman process.**

*Index Terms*—alarm systems, clustering, graph embedding, correlated alarms

## I. Introduction

Today's modern industries are comprised of thousands of sensors, actuators, and many control loops. The number of measured variables increased significantly after the pervasiveness of industrial automation like DCS in modern industries. Accessibility of a large number of measured variables increased the number of alarms in the alarm system. These alarms with weak configuration caused problems like nuisance alarm, alarm flood, false alarm, chattering alarm, and correlated alarm. These alarms have resulted in distrust of the operator in the alarm system, which might cause widespread damages.

An ideal alarm system is a system that generates one and only one alarm for each fault, and this alarm is on until the fault is resolved and confirmed by the operator. Achieving this ideal is practically difficult and impossible. Thus, an ideal alarm system is replaced by the desired alarm system. The desired alarm system is a system that realizes the instructions and requirements suggested in EEMUA-191 [1] and ANSI/ISA-18.2 [2] standards. This system improves the performance of the system and operator by helping the operator make a decision.

In today's industries, alarms might have different sources due to entanglement and relationships among different system sections. These highly correlated alarms communicate similar information of a process when an abnormality occurs, and they are known as related alarms [3]. In [4], correlated alarms are defined as alarms that are repeated regularly with a specific order. Detecting correlated alarms make detecting the fault source an easy and secure process. Thus, detecting related alarms is very useful. In [5], a color map has been presented to represent the similarity of alarms using the Jaccard similarity coefficient, effectively illustrating the correlation of the alarms and eliminating the nuisance alarms. Since then, methods based on artificial intelligence have been presented, which have demonstrated reasonable results. In [6], instead of using binary alarm sequences, an alarm log has been used, and it has been claimed that using an alarm log might perform better than binary sequences in finding the correlated alarms and alarms that are not actuated periodically. In this method, after preprocessing and assigning a vector to each alarm using the word2vec method, the alarms are clustered using a combination of K-means and AHC (Agglomerative Hierarchical Clustering) methods. Then, the clustering results are validated using the MDS (Multi-Dimensional Scaling) method; finally, a case study is presented.

An action that might help the operator make a decision is to predict the next alarm. The authors of [7] have tried to predict the next alarm using the Word2vec method and the recursive neural networks (RNNs).

Among other methods with reasonable efficiency in improving the alarm systems, data mining can be mentioned. In these methods, it is tried to extract the relationship among the alarms by finding the pattern of the alarm data. In [8], a multi-step method has been presented for alarm data preprocessing that focuses on chattering alarms and resolving the lost data problem. Using this preprocessing method, the data mining methods obtain better results.

In [9], chattering and false alarms have been studied. In this paper, instead of using a delay timer and other methods, a method based on artificial intelligence has been used to optimize the information of the alarm system. Also, it has been tried to examine fault detection. This paper has employed three steps: data collection and preprocessing, a decision tree, and fault detection. Although the presented methods have obtained acceptable results, the methods based on artificial intelligence used to find the related alarms do not consider the relationship between the alarms and the alarm log well. This paper tries to solve this problem by forming a weighted graph and using graph embedding methods. The rest of this paper is organized as follows. First, alarm systems are introduced.

Then, a method to graph alarm data is presented and graph embedding methods are reviewed. In Section VII, data clustering is described. Finally, in Section IX, a case study is presented.

## II. Alarm Data

In alarm systems, usually, the process alarms are defined such that if the variable of interest exceeds the defined threshold, the alarm is actuated. Considering this definition, problems like nuisance alarms and false alarms are created that are solved using dead-band, time delay, and filtering methods [10].

An industrial automation system usually stores the alarm data, and the previous information of the system is available. Samples of stored data alarm of a process are given in Table I. Alarm data usually indicates the alarm time, the name defined for the alarm, alarm priority, and location.

## III. The Proposed Method

The method proposed in this study is comprised of four general steps. In the first step, the data collected by the alarm system enters the preprocessing section, and the required actions are applied to it. In the second step, the weighted graph of the preprocessed data is formed. In the third step, the weighted graph is transformed to unique vectors for each alarm using the Node2Vec algorithm. In the fourth step, these vectors are clustered using the proposed clustering method; finally, the results are validated after clustering and plotting the dendrogram. First, the preprocessing step is applied to the raw alarm data. To this end, the chattering alarms are eliminated, and alarm sequences are constituted. Next, considering the presence of each alarm in the sequence, a weighted graph is formed. Then, the graph embedding is applied to the graph generated in the previous step, and each graph is converted to a vector in the embedding space. Then, the vectors corresponding to each alarm are clustered using the mentioned clustering method, and the resulting dendrogram is plotted. In the end, the results are validated by the PCA (Principal Component Analysis) method.

## IV. Preprocessing

In industrial processes, usually, the information of alarms is stored for several months or years. An alarm is a warning that informs the operator about an abnormal condition in the actuation system. A large number of these alarms form chattering alarms. Alarms that are repeated more than three times in one minute are considered chattering alarms [2]. Preprocessing comprises various steps, where eliminating the chattering alarms is one of its most important steps [8]. To eliminate these alarms, a one-minute sliding time window is considered. If an alarm is repeated more than once in this window, only the first alarm is accepted, and other alarms repeated in this window are eliminated. In the current study, it is required to form the alarm sequences. The alarm sequences are comprised of several alarms whose time interval is shorter than the defined threshold. If two alarms have a long time interval with each other, they will not have a significant relationship. Therefore, if the time interval between two alarms is larger than the defined threshold, the second alarm is considered as the first term of the next sequence. Finally, the sequences that are less than 5 are eliminated because they do not contain useful information.

## V. Graph Formation

By forming a weighted graph, it is tried to consider the relationship between the alarms. For each unique alarm, a node is considered in the graph. In the next step, and $M \times 1$ matrix is constituted for each node, where $M$ is the number of sequences. In the $W_p$ matrix, the $i^{th}$ element is 1 if the alarm $p$ exists in the sequence $i$; otherwise, it would be zero. Finally, the weight of the edge between nodes $p$ and $q$ would be:

$$e_{pq} = \frac{\sum_{i=1}^{M}(W_{p_i} \& W_{q_i})}{M} \quad (1)$$

Therefore, the number of nodes is equal to the number of unique alarms, and the matrix $e$ represents the relationship of these alarms with each other.

## VI. Embedding

Today, there exists a large volume of textual information that can be used for different purposes. The challenge is that mathematical operations cannot be applied to this data. Therefore, modern artificial intelligence methods cannot also be used. The solution is to define a vector for each word, which is called word embedding. There are various methods for this purpose that have advantages and disadvantages; one of the most popular methods is the Word2Vec method.

### A. Word2Vec Method

The Word2Vec method was first proposed in 2013 [11]. In summary, this method provides the possibility to convert the words to vectors and apply mathematical operations to these words. This method is a bilayer neural network that its input is a set of words, and its output is a vector corresponding to each word. Two structures used in this method include:

- CBOW (Continuous Bag of words)
- Skip-Gram

The architecture of the Word2Vec method is shown in Fig. 1. This method is the first word embedding method based on neural networks, and various methods have been presented after this method. One of the problems of this method is the inability to define a vector for a word that does not exist in the corpus, and one vector is defined for two words with the same writing and different meanings. In the graph embedding methods, the Skip-Gram structure is used with a random walk.

### B. Graph Embedding Method

The graph data structures have recently attracted attention. Among its applications, the recommender system used by Facebook or identifying extremist groups by Twitter can be mentioned. The graph embedding method aims to introduce a vector in $R^n$ space for each node of the graph in which n is the dimension of the embedding space, which is determined considering the number of input data and personal viewpoint. The graph embedding methods based on deep learning that is used in this paper are as follows:

- Deepwalk
- Node2Vec
- SDNE (Structural Deep Network Embedding) Which

are described briefly in the following.

TABLE I
A sample of alarm data of a process

| Alarm_tag | Subblock | Triggered Time | Finished Time | Alarm Duration | Priority |
|---|---|---|---|---|---|
| Compressor Work | Compressor & Purge | 3/7/2019 12:06:00 AM | 3/7/2019 5:00:00 AM | 294 | Low |
| Reactor Coolant Temp | Reactor | 3/7/2019 12:07:07 AM | 3/7/2019 3:06:20 AM | 179.22 | High |
| Stripper Pressure | Stripper | 3/7/2019 12:09:45 AM | 3/7/2019 1:29:56 AM | 80.19 | High |
| Product Sep Pressure | Separator | 3/7/2019 12:10:03 AM | 3/7/2019 1:29:22 AM | 79.32 | High |
| Component C to Reactor | Feed | 3/7/2019 12:18:31 AM | 3/7/2019 5:00:00 AM | 281.49 | High |

## C. Deepwalk

The Deepwalk method was first presented in 2014 [12]. This method is one of the first graph embedding methods based on deep learning that employs random walk. In the Deepwalk method, the sentences whose words constitute the graph nodes are randomly created by walking on the nodes and edges. Then, the sentences are given as input to the Word2Vec method. Implementation of the random walk method is shown in Fig. 2. In this method, the random walk is implemented for each node of the graph repeatedly to a specific number of walks. These sequences that are the graph nodes are given to the Word2Vec method. Then, the Word2Vec method creates vectors in the specified embedding space, which can be used in machine learning algorithms.

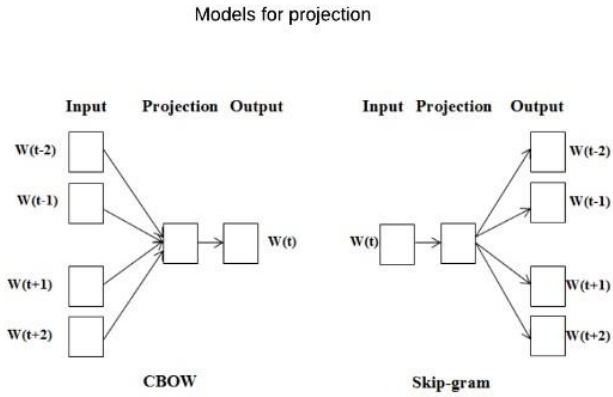

Fig. 1. The architecture of the Word2Vec method; CBOW and Skip-Gram structures [11].

## D. Node2Vec Method

The Node2Vec method was presented in 2016 to enhance the Deepwalk method [13]. This method tries to cover the shortcomings of the Deepwalk method. That is, in the Deepwalk method, movement from one node to another is random, so one path might be repeated several times, or a part of the graph might not be accessed. In the Node2Vec method, two constants, $p$, and $q$ are defined to avoid repetition, and the random walks remain around the origin node, and the nodes linked to the node of interest attract more attention. The architecture of the Node2Vec method is shown in Fig. 3. Also, $\alpha$ is obtained as follows:

$$\alpha_{\{pq\}(t,x)} = \begin{cases} \frac{1}{p} & if \ d_{tx} = 0 \\ 1 & if \ d_{tx} = 1 \\ \frac{1}{q} & if \ d_{tx} = 2 \end{cases} \quad (2)$$

where $d_{tx}$ is the number of closest edges of the path between nodes $t$ and $x$, after creating the sentences, other steps are the same as the Deepwalk method. In this paper, the Node2Vec method is used to create the corresponding vectors of each node. In the following, the generated vectors are clustered.

## VII. Clustering

To help the operator make a faster and more efficient decision, the operator know of the process and the alarm system. Therefore, clustering is performed to help the operator better recognize the alarm system and understand the relationships among the alarms. Since the initial selection of the centers of the clusters affects the output of the K-means clustering significantly, aggregated clustering is used [6], which combines the K-means and AHC algorithms. The K-means algorithm is executed $r$ times, and the output is copied in an $H \times k$ dimensional $R^{(r)}$ matrix. In this matrix, the $R_{ih}^{(r)}$ element is 1 if the $i^{th}$ data is assigned cluster h in the $r^{th}$ execution; otherwise, it is zero. By combining $M$ solutions and their matrices, a matrix in the following form is formed.

$$R = [R^{(1)}, R^{(2)},...,R^M] \quad (3)$$

The aggregated distance D is defined as follows [6].

$$D = 1 - \frac{1}{M} R R^T \quad (4)$$

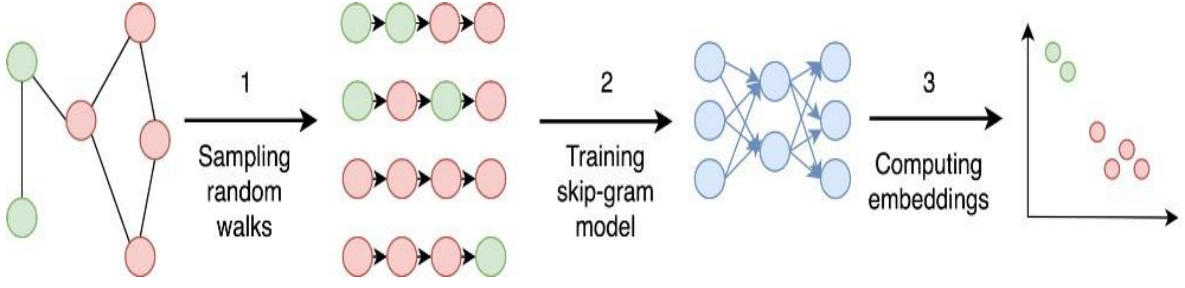

Fig. 2. An example of applying the random walk to a graph [12].

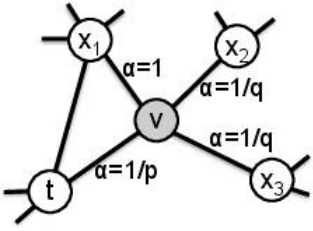

Fig. 3. The architecture of the Node2Vec network considering the parameters $p$ and $q$ [13].

where 1 is a matrix whose elements are 1. The matrix $D$ represents a measure of dissimilarity that its input might be the AHC method. To find the best value of $K$, the algorithm is executed for 2 to $k'_{max}$. Then, $k'_{max}$ is obtained by comparing each execution with the previous execution. By increasing $k'_{max}$, the K-means method is executed for random initial values for $M_{runs}$ times. After obtaining these values, the aggregated distance matrix $D_{(k'_{max})}$ is calculated considering (3) and (4) and considering $M = (k'_{max} - 1)M_{runs}$. To create a termination measure, the following measure, which is defined as the sum of square error, can be used.

$$\varepsilon(k'_{max}) = \sum_{i<j}\left[D_{ij}(k'_{max} + 1) - D_{ij}(k'_{max})\right]^2$$

The parameter $k_{max}$ is the smallest $k'_{max}$ that converges $\varepsilon(k'_{max})$. Then, the input of the converged $D_{(K_{max})}$ would be the AHC method. Considering the clustering carried out by the AHC method, dendrograms of the vectors are plotted and analyzed. These dendrograms can help operators to manage alarm systems functionally and avoid making unformed decisions.

## VIII. Validation

In many studies, it is essential to visualize data and evaluate the results. However, since data is usually in a space larger than $R^3$, visualization is not possible. Thus, the data should be visualized in lower dimensions. Therefore, dimension reduction algorithms are used. Alarm data are also in high dimensions because of numerous features. In this study, the PCA method is used for dimensionality reduction and helps to visualize alarm data [14].

This algorithm converts n-dimensional data to k-dimensional data using singular values. The steps of this algorithm are given in the following.

1) Standardizing raw data: the input data of this method should have a zero mean. Therefore, the data mean should be calculated and subtracted from the data.
2) Calculating the covariance matrix: this matrix is obtained using the following equation.

$$\Sigma = \frac{1}{m}\sum_i^m (x_i)(x_i)^\top, \Sigma \in R^{n \times n} \quad (6)$$

where $x$ is the data matrix in this case embedded alarm data and $\Sigma$ is the covariance matrix, and $m$ is the number of data.

3) Calculating the eigenvector and the eigenvalues: in the third step, the eigenvector and eigenvalue of the covariance matrix are calculated, and the eigenvector matrix is:

$$U = \begin{bmatrix} \vdots & \vdots & & \vdots \\ u_1 & u_2 & \ldots & u_3 \\ \vdots & \vdots & & \vdots \end{bmatrix}, u_i \in R^n \quad (7)$$

where, is the eigenvector and $U$ is the eigenvector matrix.

4) Calculating the vectors in a lower dimension space: after calculating the eigenvector matrix $U$, depending on $k$ (which is the dimension of the new space), the first $k$ vectors of the eigenvector matrix are sorted based on the eigenvalue are selected. Then, the initial vectors are multiplied by the selected eigenvector matrix, and new vectors are generated.

$$x_i^{new} = \begin{bmatrix} u_1^T x_i \\ u_2^T x_i \\ \vdots \\ u_n^T x_i \end{bmatrix} \quad (8)$$

Thus, the vectors in the $n$-dimensional space are converted into vectors in the $k$-dimensional space, where $k$ is smaller than $n$.

## IX. Case Study

In this section, using the alarm data of the Tennessee-Eastman process and the graph embedding and clustering methods, it is tried to find similar alarms and detect the source of correlated alarms. It is assumed that the alarm system is designed based on correct principles [15]. The studied scenarios are given in Table II. In this case study, 4000 initial alarms of each fault are selected. The number of triggers of each alarm is shown in Fig. 4. Also, a tag corresponding to each fault with equal intervals is included in the alarm data. This is done to help fault detection after clustering and relating the alarms with the faults.

After preprocessing and eliminating the chattering alarms, the alarms that are repeated more than three times in one minute, a 5-minute sliding window is applied to create alarm sequences. Then, the weighted graph is constituted. Next, the Node2Vec method is applied to the matrix. The variables measured in the Tennessee-Eastman process and the tags corresponding to the alarms are shown in Table III.

### A. Graph Embedding

After preprocessing and constituting the graph, the weighted graph is given to the Node2Vec method as input, and a vector is assigned to each alarm. The dimension of the embedding space is considered 300. Using the cosine similarity measure, the similarity of the vectors corresponding to each alarm is obtained and visualized in Fig. 5 using a heat Map. As can be seen, the brighter is the color of the overlapping blocks of two alarms, and the alarms are more similar regarding the cosine similarity measure.

### B. Clustering of Tennessee-Eastman process alarms

Aggregated clustering is used for data clustering. First, the number of clusters in the K-means method is selected as 4 using the elbow method. Then, this method is applied 100 times to the data with random cluster centers ($M = 100$) and stored in the matrix $R$. Then, the matrix $D$ is formed using (4), and the dissimilarity matrix is clustered using the AHC clustering method, and its dendrogram is plotted (Fig. 6). As shown in Fig. 6, the data combined at low levels promise a close and relationship with each other. For example, RT↑ (Reactor

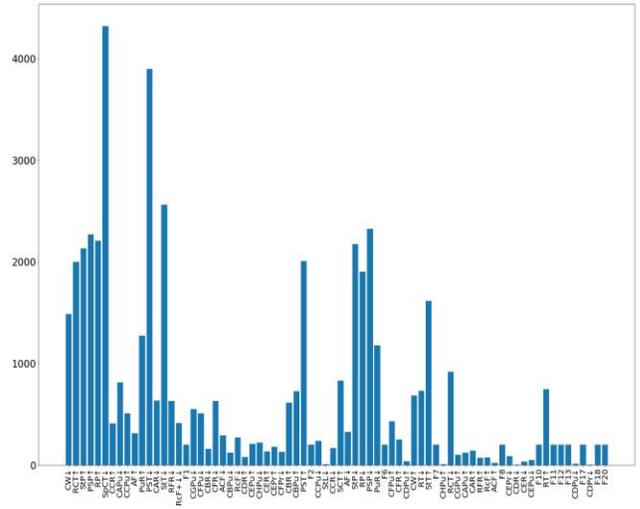

Fig. 4. The number of triggers of each alarm in the case study.

TABLE III
The variables of the Tennessee-Eastman process and the corresponding alarm tags.

| Name | Process Variable | Unit | Name | Process Variable | Unit |
|---|---|---|---|---|---|
| AF | A feed (stream 1) | $kscmh$ | CAR | Component A to Reactor | $Mole\%$ |
| DF | D feed (stream 2) | $kg/h$ | CBR | Component B to Reactor | $Mole\%$ |
| EF | E feed (stream 2) | $kg/h$ | CCR | Component C to Reactor | $Mole\%$ |
| ACF | A & C feed (stream 4) | $kscmh$ | CDR | Component D to Reactor | $Mole\%$ |
| RcF | Recycle flow (stream 8) | $kscmh$ | CER | Component E to Reactor | $Mole\%$ |
| RFR | Reactor feed rate (stream 6) | $kscmh$ | CFR | Component F to Reactor | $Mole\%$ |
| RP | Reactor pressure | $kPagauge$ | CAPu | Component A in Purge | $Mole\%$ |
| RL | Reactor level | $\%$ | CBPu | Component B in Purge | $Mole\%$ |
| RT | Reactor temperature | $°C$ | CCPu | Component C in Purge | $Mole\%$ |
| PuR | Purge Rate | $kscmh$ | CDPu | Component D in Purge | $Mole\%$ |
| PST | Product Sep Temp | $°C$ | CEPu | Component E in Purge | $Mole\%$ |
| PSL | Product Sep Level | $\%$ | CFPu | Component F in Purge | $Mole\%$ |
| PSP | Product Sep Pressure | $kPagauge$ | CGPu | Component G in Purge | $Mole\%$ |
| PSU | Product Sep Underflow | $m^3/hr$ | CHPu | Component H in Purge | $Mole\%$ |
| StL | Stripper Level | $\%$ | CDPr | Component D in Product | $Mole\%$ |
| StP | Stripper Pressure | $kPagauge$ | CEPr | Component E in Product | $Mole\%$ |
| StU | Stripper Underflow | $m^3/hr$ | CFPr | Component F in Product | $Mole\%$ |
| StT | Stripper Temp | $°C$ | CGPr | Component G in Product | $Mole\%$ |
| StSF | Stripper Steam Flow | $kg/h$ | CHPr | Component H in Product | $Mole\%$ |
| CW | Compressor Work | $kW$ | | | |
| RCT | Reactor Coolant Temp | $°C$ | | | |
| SpCT | Separator Coolant Temp | $°C$ | | | |

TABLE II
The scenarios studied in the Tennessee-Eastman process.

| Fault Scenario | Type | Fault Description |
|---|---|---|
| F1 | Step | A/C-ratio of stream 4, B composition constant |
| F2 | Step | B composition of stream, A/C-ratio constant |
| F6 | Step | A feed loss (stream 1) |
| F8 | Step | C header pressure loss (stream 4) |
| F10 | Random | C feed (stream 4) temperature |
| F11 | Random | Cooling water inlet temperature of reactor |
| F12 | Random | Cooling water inlet temperature of separator |
| F13 | Drift | Reaction kinetics |
| F17 | Random | Deviations of heat transfer within reactor |
| F18 | Random | Deviation of heat transfer within condenser |
| F20 | Random | (Unknown) |

Temperature High) and RP↑ (Reactor Pressure High) are combined at a low level, indicating the relationship of these two alarms, which is verified by the process logic also; because increasing the reactor temperature increases the pressure. These results can help the operator to make better and faster decisions when required. For example, if the RT↑ alarm is triggered, the operator can predict possible actuation of RP↑, and if this alarm is triggered, the operator can take a more centralized action considering the dendrogram and knowing the relationship of these two alarms with each other, and react to the alarms more systematically. In another example, to validate the results, as shown in Fig. 6, fault 12 and Spt (Separator Coolant Temperature High) alarm are combined at a low level and are related together according to the process logic, and fault 12 might be the cause of alarm

$$x_i^{new} = \begin{bmatrix} u_1^T x_i \\ u_2^T x_i \\ \vdots \\ u_n^T x_i \end{bmatrix} \quad (8)$$

Spt. Thus, the operator can detect the cause of actuating this alarm and resolve it quickly.

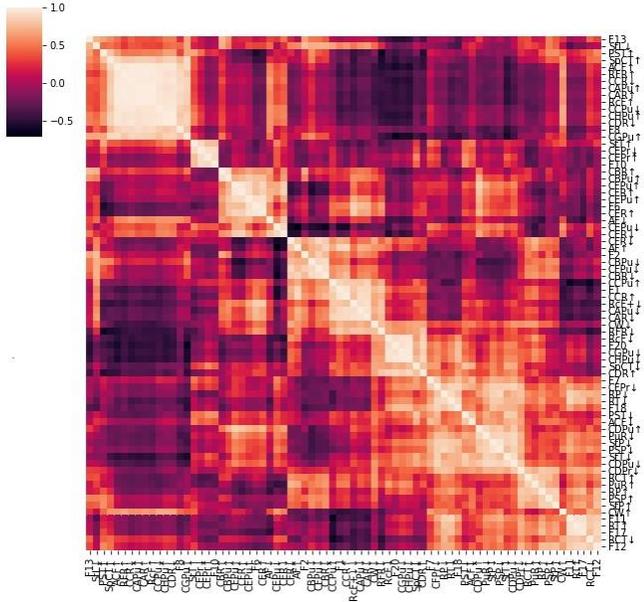

Fig. 5. Heat map of the Tennessee-Eastman process alarms.

Considering the alarms received by the operator that exceed the responsiveness capacity of the operator, the proposed method can reduce the volume of input alarms by eliminating the chattering alarms, finding similar alarms, and predicting alarms that will be triggered in the future. Also, it helps the operator make better decisions by finding the relationships between the alarms, preventing distrust of the operator in the alarm system and delivering a more favorable alarm system to the operator.

C. Validating the Tennessee-Eastman Process Clustering

To validate and observe the clustering results, clustered vectors should be plotted, and the clustering should be validated. The output vectors of the Node2Vec method are in the $R^{128}$ space. To validate the clustering, this data is first converted to 2-dimensional data using PCA in Python using the Scikit-Learn library. As shown in Fig. 7, the data is divided into four clusters represented in blue, green, yellow, and purple. As can be seen, the four clusters are separated reasonably. Therefore, it can be said that clustering performed favorably, and the operator can prevent possible threats relying on the performed clustering and using the dendrogram.

By ensuring clustering of the Tennessee-Eastman process alarms, the proposed method can be used in more complex systems with more alarms. The alarm systems are usually made of hundreds or thousands of alarms; thus, an embedding space of 128 is proper.

In [6] and [7], the Word2Vec method is used for clustering alarm data. In this method, the alarm system data is converted to alarm sequences using a sliding window. Then, the sequences are used as input sentences of the Word2Vec method. Then, the output vectors corresponding to each alarm are clustered using cumulative clustering, which combines K-means and AHC clustering methods. Next, the output vectors of the Word2Vec method are converted to vectors in lower dimensional space using the PCA method, and the result is plotted. These vectors are shown in Fig. 8.

Although the relationship between alarms of a sequence is considered in the Word2Vec method, the relationship of the sequences is not calculated well. Also, this method is more vulnerable to chattering alarms; that is, repetition of an alarm in one sequence might significantly affect the output result of the Word2Vec method. Also, in this method, the clusters are entangled, and it is difficult to separate them. But in the proposed method, while constructing the graph, the presence or absence of the alarms in the sequence is checked, making it robust against chattering alarms. Also, due to the construction of the weighted graph, the relationship between alarms and sequences is modeled appropriately and Node2Vec method Turns this relation into their corresponding vectors. Given that the relationship between the alarms is well-considered, clustering also shows the correlated alarms well, and identifying these alarms will be of great help to the operator.

## X. Conclusion

One of the most challenging problems of alarm systems is that the number of incoming alarms exceeding the standard, resulting in operator confusion and inadequate response to triggered alarms. The purpose of this article is to provide a way to help the operator make better and more effective decisions. The process of this method involves constructing a weighted matrix of alarms stored in the alarm log using their relationships and using the Node2Vec method to embed vectors corresponding to existing alarms and then clustering these vectors using clustering methods K-means and AHC. Tennessee-Eastman process alarms has since been conducted. In this case study, after clustering and drawing the dendrogram, the combination of low-level alarms indicates a relationship between them, A case study on the which is confirmed by process logic. By eliminating chattering alarms in the preprocessing phase, as well as extracting correlated alarms and predicting alarms, this method can help the operator make faster and more systematic decisions, which avoids financial and human risks. This method can also be used to identify the source of the defect, which is the subject of further studies.

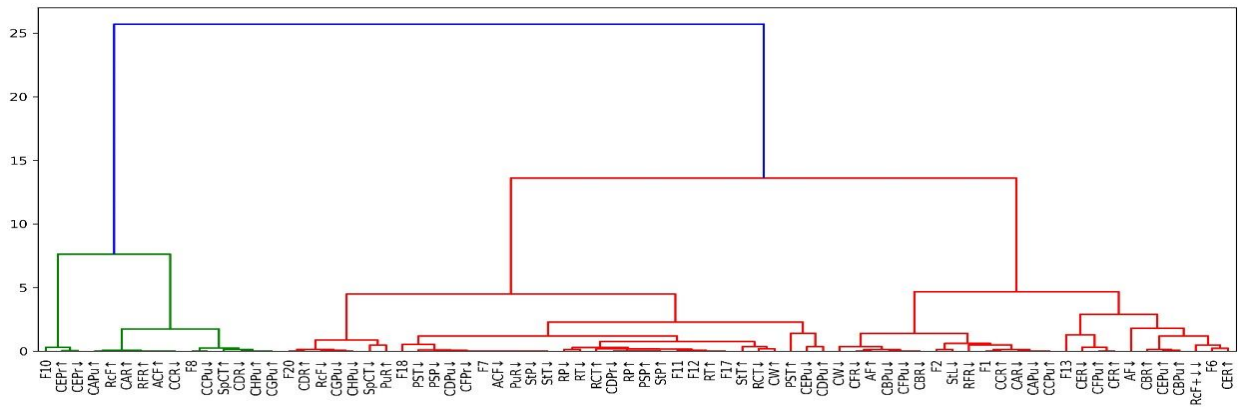

Fig. 6. Dendrogram of the vectors corresponding to alarms of the Tennessee-Eastman process.

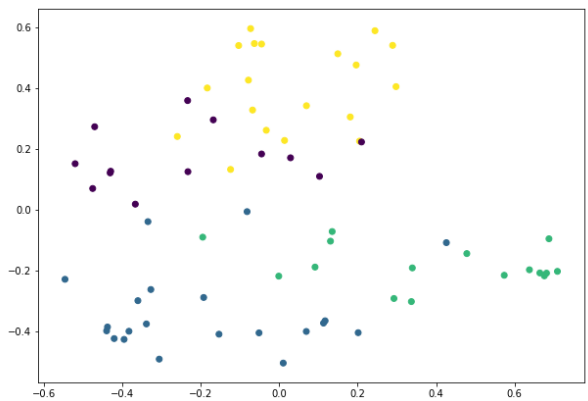

Fig. 7. The vectors corresponding to alarms of the Tennessee-Eastman process are plotted in $R^2$ space

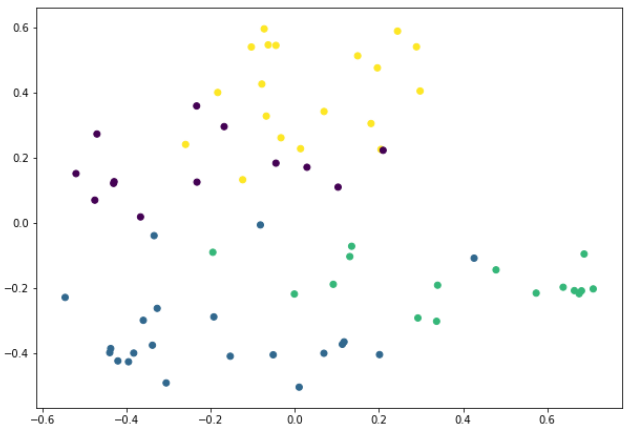

Fig. 8. The vectors corresponding to alarms of the Tennessee-Eastman process are embedded using the Word2Vec method.